\newif\ifreview
\reviewtrue

\documentclass{article}
\usepackage{ijcai26}

\usepackage{times} % Use the postscript times font!
\usepackage{soul}
\usepackage{url}
\usepackage[hidelinks]{hyperref}
\usepackage[utf8]{inputenc}
\usepackage[small]{caption}
\usepackage{graphicx}
\usepackage{amsmath}
\usepackage{amsthm}
\usepackage{amssymb}
\usepackage{dsfont}
\usepackage{booktabs}
\usepackage{algorithm}
\usepackage{algorithmic}
\usepackage[switch]{lineno}
\usepackage{xcolor}

\title{CURED: Creating, Understanding, and Repairing Errors Demonstrator}
\author{
Nicholas Chandler$^1$\footnote{Contact Author}\and
Sebastian Jäger$^1$\and
Philipp Jung$^1$\And
Felix Bießmann$^{1,2}$\\
\affiliations
$^1$Berliner Hochschule für Technik\\
$^2$Einstein Center Digital Future\\
\emails
\{nich1834, Sebastian.Jaeger, Philipp.Jung, Felix.Biessmann\}@bht-berlin.de,
}

\begin{document}
\maketitle

\begin{abstract}
Detecting and cleaning errors in tabular data is a prerequisite for data intense software applications. Recent research at the intersection of Machine Learning (ML) and Database Management Systems (DBMS) highlights the potential of statistical learning algorithms for error detection and cleaning. This paper combines our recent work on ML-based data cleaning and error models in a unified demonstrator. The web application allows users to upload tabular data, perturb the data with realistic data dependent errors and use modern ML methods to clean and understand error mechanisms in data. Our demonstrator\footnote{Video: \scriptsize{\url{https://cloud.bht-berlin.de/s/6m8twqEcsSZKiHT}}} helps to bridge the gap between theoretical advancements and intuitive practical insights in the context of error models and data cleaning algorithms for tabular data. The demonstrator is available at \scriptsize{\url{https://cured.demo.calgo-lab.de/}}.
% demonstrates how errors in tabular data can be modeled, mitigated, and investigated. The demo begins with the upload of a dataset, then a specific error model can be chosen to generate errors in the data using \texttt{tab-err}. Thereafter, the data are cleaned with \emph{Conformal Data Cleaning}, where the impact of the cleaning on a downstream Machine Learning (ML) task is examined. Finally, the table is characterized with respect to the distribution of the errors in various columns, the \emph{error mechanism}, using \emph{MechDetect}. 
% The demonstration bridges the theoretical aspects of these three works with a hands-on tool for experimentation on novel datasets as well as code samples for those interested in utilizing the methods elsewhere.
\end{abstract}

\section{Introduction} \label{sec:intro}
Tabular data is prevalent in industry and research due to compatibility with the relational model, spreadsheets, statistical, and ML algorithms~\cite{silberschatz2019database,jiang2025tabularsurvey}. Errors in tabular data remain a substantial problem across a variety of applications~\cite{powell2009impactOfErrors,gurupur2025EHRerrors} and provide a rich field of research~\cite{abedjan2016detecting,jagerBenchmarkDataImputation2021}. With the advent of more data intense software applications, there are also novel regulatory frameworks and standards related to data quality that companies must adhere to~\cite{ISOIEC25024}. Doing so will require realistic error models and a high degree of automation for data cleaning methods~\cite{schelterAutomatingLargeScaleData2018,biessmann2021automated}. 

While recent research has advanced the fidelity at which error models for tabular data can represent realistic errors~\cite{jung2025realistic}, they are often difficult to set up and test because of the large number of methods available for cleaning data, detecting errors and characterizing error mechanisms. Moreover, we believe that demonstrators can help build intuition for the theoretical contributions in the context of data quality at the intersection of Machine Learning (ML) and Database Management Systems (DBMS). Some popular open source tools such as OpenRefine~\cite{openrefine} were great examples of how important interactive demonstrators of data cleaning tools can be, but the implementation of OpenRefine does not integrate modern data cleaning software. We hence provide the following demonstration as a means for users to explore several facets of modern ML based tabular error research in one place.
This demo primarily bridges the theoretical and empirical results of research on generation of realistic data dependent errors~\cite{jung2025realistic}, work on conformalized data cleaning~\cite{jager2024conformal}, and novel approaches to detect realistic data dependent error generation mechanisms~\cite{jung2025mechdetect}. 

\begin{figure}
    \centering
    \includegraphics[width=1.0\linewidth]{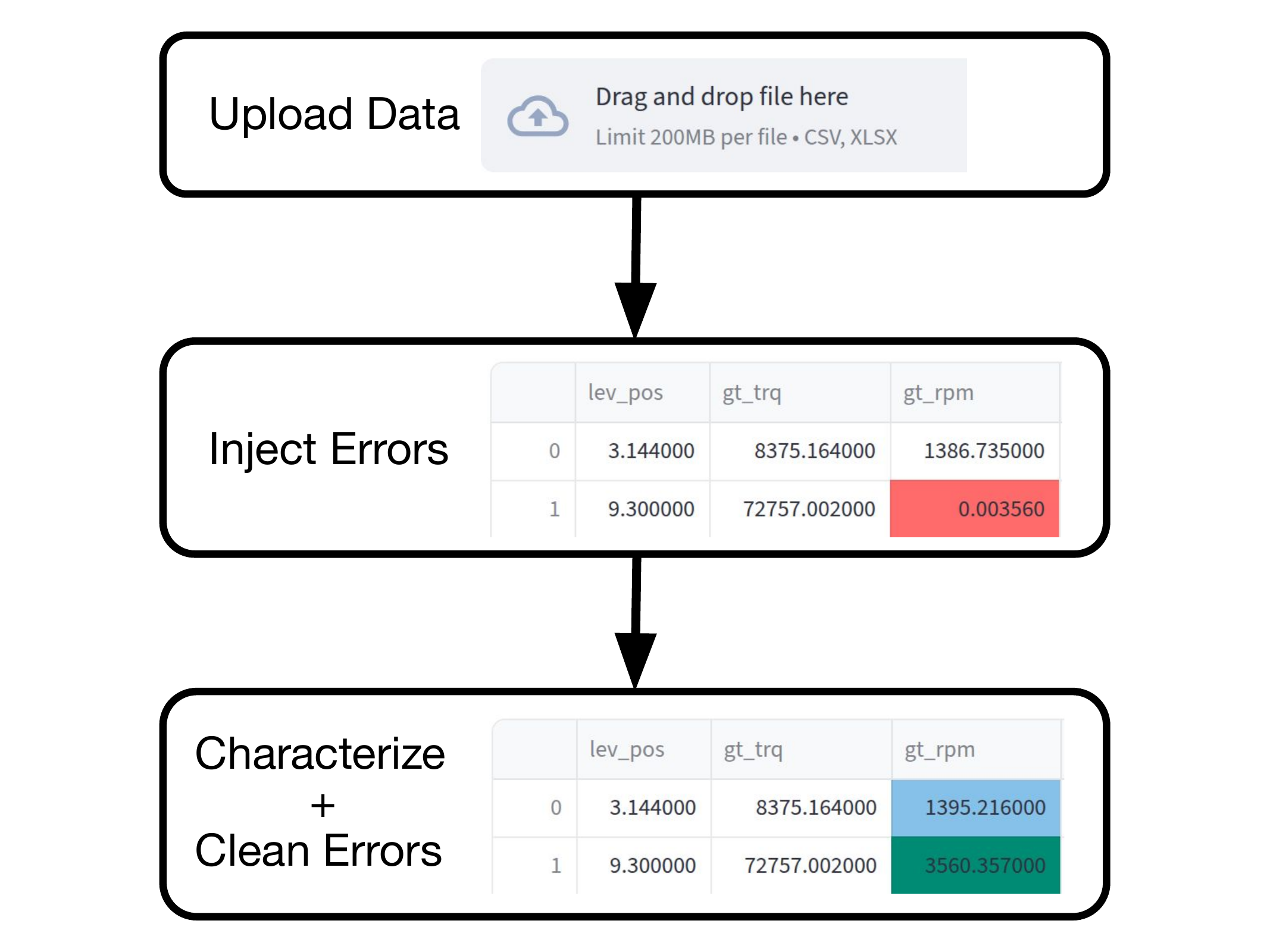}
    \caption{Overview of CURED. Users can upload data, perturb data with predefined realistic error types and then use ML tooling to detect, characterize and clean errors.}
    % Link to figure https://docs.google.com/presentation/d/1gqZhGZiKrCBpFTG1eqmBfqbVb5kz4cvajcIGWxSaXu0/edit?usp=sharing
    \label{fig:flowchart}
\end{figure}

In \autoref{fig:flowchart} we show an overview of the demonstrator's usage. After uploading a data set or loading a predefined one, realistic, data-dependent errors are injected using \texttt{tab-err}~\cite{tab_err}, an error generation library that decouples the error mechanism and the error type, thus allowing fine-grained calibration and definition of error scenarios~\cite{jung2025realistic}. 
Having perturbed the data this way, users can leverage ML-based data cleaning methods, here Conformal Data Cleaning (CDC)~\cite{jager2024conformal}, to clean the data and evaluate the result on a downstream ML task. CDC utilizes conformal prediction to locate and correct erroneous cells with user-defined hyperparameters.
Finally, the recently developed \texttt{MechDetect}~\cite{jung2025mechdetect} can be used to further investigate the error generating mechanism and categorize the error mechanisms used to perturb the data.
%to the detection and correction of single errors, a recently developed algorithm for categorizing error mechanisms can be used to further investigate the error generating mechanism.
\texttt{MechDetect} seeks to characterize data-dependent errors by comparing the performance of ML models on predicting an error mask, yielding the error mechanism affecting the column of interest.

These three components: error injection, error characterization, and error correction are combined into one demonstrator which enables users to develop an intuition for errors, and error detection, and error cleaning algorithms in tabular data. All code is released under permissive open source licenses\footnote{https://github.com/calgo-lab} and can be effectively integrated into existing Python workflows. Although we restrict the algorithms in the backend to a small set of ML-based methods, the demonstrator lends itself easily to extensions with other error generators or data cleaning approaches.

\section{CURED Components} \label{sec:demo}
The following sections highlight first the essential parts of the demonstrator (\autoref{fig:flowchart}), before explaining the backend components in more detail and providing empirical results that document their performances quantitatively. The demonstrator can be accessed at \url{https://cured.demo.calgo-lab.de/}.

%Looking closer at the components in \autoref{fig:flowchart}, after uploading or selecting a tabular data set, users can customize error models used to perturb their data. Next, conformal data cleaning finds and corrects errors. Finally, users can characterize the introduced errors using \texttt{MechDetect} to determine the error mechanism of a selected column.

\subsection{Loading Data} \label{sec:data-ingest}
The demonstrator first loads a dataset. Users can either choose the preloaded dataset, a subset consisting of eight columns of the nuclear power plant dataset with OpenML ID 44969~\cite{bischl2021openml}. %A subset of 8 columns with abbreviated feature names is used for compactness.
Or they upload datasets in the formats csv or xlsx. The requirements are that the dataset has fewer than 10 columns, fewer than 10,000 rows, and that it has a column named \emph{target}.
Additionally, 80\% of the dataset will be used for training the conformal cleaner and downstream ML model, thereby leaving 20\% for use in the error generation, cleaning, and characterization.

% \begin{figure}[htbp]
%     \centering
%     \includegraphics{figures/data-loader.png}
%     \caption{Data ingestion phase of the demo}
%     \label{fig:data_ingestion}
% \end{figure}

\subsection{Creating Errors} \label{sec:generate-errors}
\begin{figure}[htbp!]
    \centering
    \includegraphics{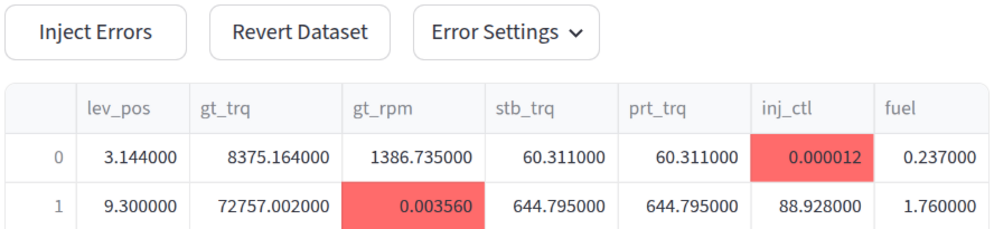}
    \caption{Error generation phase of the demo}
    \label{fig:errors}
\end{figure}

The user can click \emph{Inject Errors} to perturb their data, as shown in Figure~\ref{fig:errors}. Custom error scenarios can be configured in the \emph{Error Settings}. Additionally, the dataset can be replaced with the originally uploaded data by clicking \emph{Revert Dataset}. The result of this stage is a table with perturbed cells highlighted in red.

Errors are generated using the package \texttt{tab-err}~\cite{tab_err}. Following the error generation model in \cite{jung2025realistic} errors are parametrized by error type, error mechanism, and error rate. 
The error type is simply how the cell in the table is actually altered. For example a numeric cell may be multiplied by 10 if the wrong unit was recorded. A categorical cell may have an 'h' character replaced by a 'j' character due to the proximity of the keys on the qwerty keyboard and a clumsy typist.
The error rate is the proportion of cells in the table that contain errors.
The error mechanism describes the distribution of error locations in a table. 
Error mechanisms are exhaustively categorized by Errors Completely At Random (ECAR), Errors At Random (EAR), and Errors Not At Random (ENAR) as shown in source~\cite{jung2025realistic}. The dependency structure of error mechanisms are summarized in Table~\ref{tab:error_mechanisms}.

\begin{table}[htbp]
\centering
\caption{Error Mechanism Dependency Structure}
\label{tab:error_mechanisms}
\begin{tabular}{lll}
\toprule
\textbf{Mechanism} & \textbf{Dependence}\\
\midrule
Errors Completely At Random & Independent of Data\\
Errors At Random  & Another Column\\
Errors Not At Random & Same Column\\
\bottomrule
\end{tabular}
\end{table}

The error models used by tab-err are a generalization of Rubin's missing data theory~\cite{rubin1976inference}. Specifically, they allow general error types to be modeled where missing values are simply one of these types.

\subsection{Repairing Errors} \label{sec:clean-errors}
\begin{figure}[htbp!]
    \centering
    \includegraphics{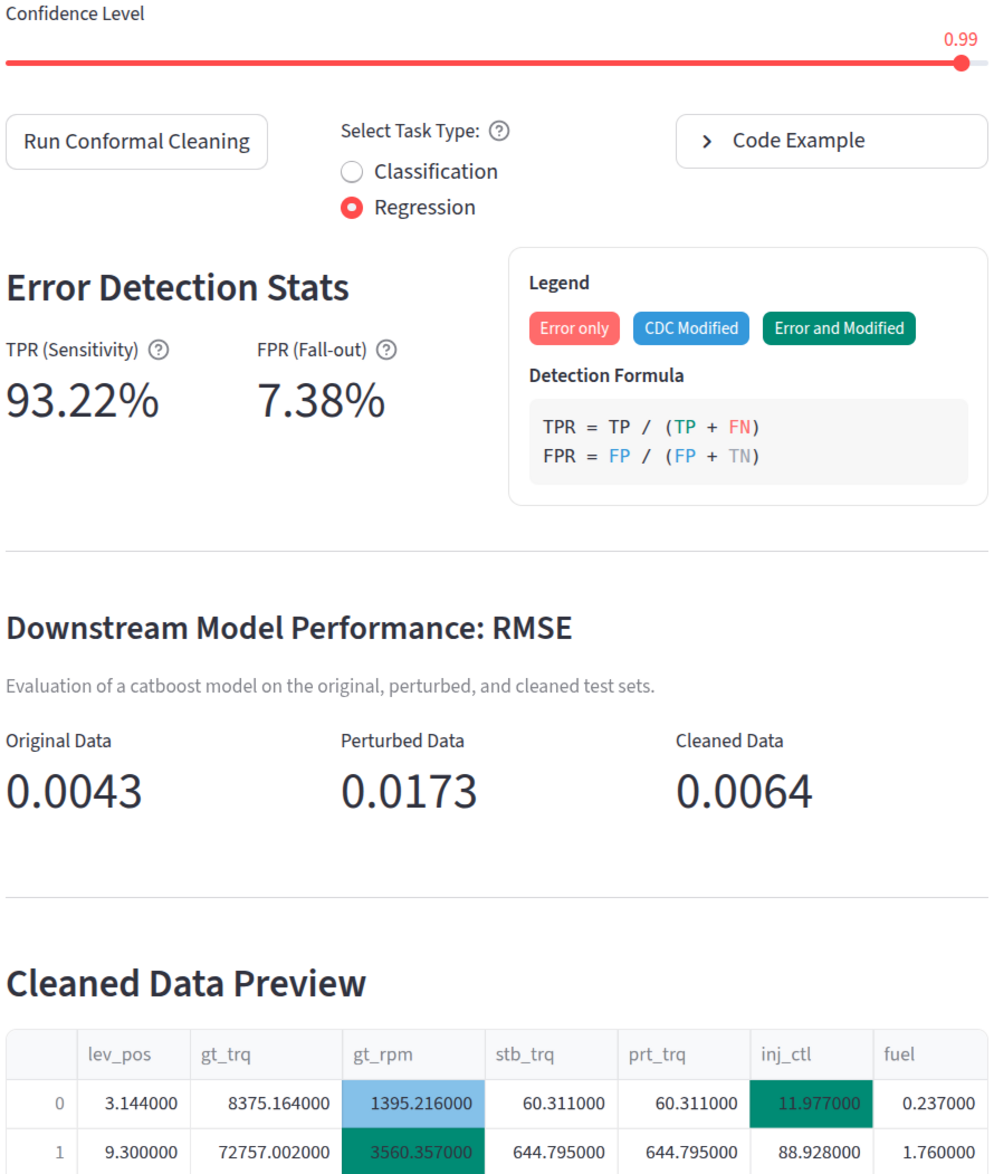}
    \caption{Error cleaning phase of the demo}
    \label{fig:cleaning}
\end{figure}
The user can click the \emph{Run Conformal Cleaning} button to run the cleaning process and downstream ML task, as shown in Figure~\ref{fig:cleaning}. The task type should be specified here by the user as well. A code example of the cleaner is provided in the \emph{Code Example} drop down of the CURED user interface. Finally, the confidence level of the cleaner (determining the conservatism of CDC) can be specified by the user.

The cleaning backend employs an ML based cleaning algorithm. In~\cite{jagerBenchmarkDataImputation2021} show that data imputation is implemented by training a tabular ML model for every column of a table, using all other columns as input features. 
Formally, $imputer$ trains $d$ models on a subset of columns, where $X_c = X_{\{ 1, \dots, d \} \setminus \{ c \} }$ are the features and $y_c = X_c$ the target to fit $imputer_c$ to impute values in column $c \in \{1, \ldots, d \}$.
Conformal Data Cleaning~\cite{jager2024conformal} generalizes this idea and uses the imputation models to detect and correct errors in tabular data by calibrating the uncertainty of each individual imputation model.
Conformal prediction is an uncertainty calibration method that turns black-box models into set predictors, which give statistically rigorous confidence sets for a user-defined miscoverage rate $\alpha$.
For a given test set $X^{test}$, these prediction sets $\mathcal{C}(X^{test})$ guarantee to contain the correct label $y^{test}$ in at least $1 - \alpha$ percent of the cases, formally $\mathbb{P}(y^{test} \in \mathcal{C}(X^{test})) \ge 1 - \alpha$.
If the cell $X^{test}_{i,c}$ exists in the prediction set $\mathcal{C}(X^{test}_{i,c})$ predicted by $cleaner_{c}$, we consider this value as correct and leave it as is.
If it does not exist in the corresponding prediction set, we assume it is incorrect.
This \emph{error detection} step computes a binary mask, which marks incorrect cells as $1$ and correct ones as $0$.
Replacing incorrect cells of the input data with \emph{NaN}s essentially casts the cleaning problem into a missing value imputation problem.
We then directly replace incorrect cells with the point predictions of the underlying ML model, which concludes the \emph{error cleaning} step.

In ~\autoref{tab:cdc-results}, we detail the error detection True Positive Rate (TPR) and False Positive Rate (FPR) as well as the Downstream Improvement (DSI)\footnote{DSI is improvement after cleaning relative to perturbed performance}.
We follow the experimental setting of the original paper, fix the confidence level to $0.999$, run the experiments over the error mechanisms (ECAR, EAR, and ENAR), and report the median results.
Note that depending on the specific setting, CDC's performance varies, for more details, we refer the reader to~\cite{jager2024conformal}.
\begin{table}[h]
    \centering
    \caption{CDC Error Detection True Positive Rate (TPR) and False Positive Rate (FPR) and Downstream Improvement (DSI) by Error Mechanism}
    \begin{tabular}{lrrr}
        \toprule
        Metric & EAR & ECAR & ENAR \\
        \midrule
        TPR & 0.6504 & 0.6488 & 0.6459 \\
        FPR & 0.0065 & 0.0066 & 0.0070 \\
        DSI & 0.0349 & 0.0069 & 0.0285 \\
        \bottomrule
    \end{tabular}
    \label{tab:cdc-results}
\end{table}

\subsection{Understanding Errors} \label{sec:characterize-errors}
Apart from the detection of individual error locations, it can be helpful to understand the error mechanisms for deciding error mitigation strategies. Using the \texttt{MechDetect} algorithm, CURED reports the predicted and injected error mechanisms. It also reports the p-values of the ML-model performance comparisons within \texttt{MechDetect}. The error mechanisms (as in Section~\ref{sec:generate-errors}) consist of Errors Completely At Random (ECAR), meaning there is no dependency of errors on data, or Errors At Random (EAR) / Errors Not At Random (ENAR), meaning that errors depend on observed data in specific columns of the data (EAR) or on unobserved other data (ENAR).

In an exemplary setting, we conducted some experiments to demonstrate the efficacy of the error mechanism detection.
The setup was the same as in the original paper~\cite{jung2025mechdetect} but we used the error types as in the demo\footnote{Numeric: \emph{Outlier}, \emph{WrongUnit}, \emph{AddDelta} \\\indent Categorical: \emph{Typo}, \emph{Replace}, \emph{Extraneous}}. Specifically, there were fewer applicable datasets for the two cases with 109 for numeric and 17 for categorical. Finally, Catboost~\cite{Prokhorenkova2018CatBoost} was applied instead of LightGBM~\cite{Ke2017LightGBM}.

% For numeric columns, we used the error types: \emph{Outlier}, \emph{WrongUnit}, and \emph{AddDelta} with default configurations on 109 applicable datasets. For categorical columns, we used the error types: \emph{Typo}, \emph{Replace}, and \emph{Extraneous} on default configurations with 17 applicable datasets. All datasets had less than 100,000 rows and 500 columns and were sourced from several tabular data benchmarks~\cite{bischl2021openml,fischer2023openmlctr,grinsztajn2022tree,jagerBenchmarkDataImputation2021}. The error rates used were 0.1, 0.25, 0.5, 0.75, and 0.9. The ML model used was CatBoost with the default hyperparameters, the performance metric of the tasks was ROC-AUC, and alpha for the internal statistical tests was 0.05.

The \emph{accuracy} of MechDetect for numeric and categorical error types under the three error mechanisms and aggregated over selected datasets, error rates in a balanced number of scenarios for each mechanism is shown in Table~\ref{tab:med-acc}. 

\begin{table}[h]
\centering
\caption{MechDetect Accuracy by Column Type and Error Mechanism}
\begin{tabular}{llll}
    \toprule
    Column Type & ECAR & EAR & ENAR \\ 
    \midrule
    Numeric & 0.9421 & 0.9900 & 0.7327 \\ 
    Categorical & 0.9295 & 0.9984 & 0.7079 \\ 
    \bottomrule
\end{tabular}
\label{tab:med-acc}
\end{table}

\section{Conclusion} \label{sec:conclusion}

This work bridges a body of research on errors in tabular data with end users, allowing them to understand ways in which errors can be modeled and mitigated.
Errors were generated using realistic models, 
thereafter they were cleaned using an ML-based method with conformal guarantees,
finally they were characterized using a novel error-mechanism detection algorithm.
Future work could explore the utilization of MechDetect in a cleaning pipeline, or the maintenance and improvement of the error types in tab-err. Finally, another direction could be the employment of methods in industrial-scale applications.

% \section{Acknowledgments}

% Felix should probably write this.

\bibliographystyle{named}
\bibliography{ijcai26}

@misc{tab_err,
  author       = {Calgo-Lab},
  title        = {tab\_err: A Python package for detecting and correcting tabular data errors},
  year         = {2026},
  publisher    = {GitHub},
  url          = {https://github.com/calgo-lab/tab_err},
  version      = {latest},
  note         = {Accessed: 2026-02-09}
}

@misc{openrefine,
  author       = {OpenRefine-Team},
  title        = {OpenRefine: A free, open source, power tool for working with messy data},
  month        = mar,
  year         = 2024,
  publisher    = {Zenodo},
  version      = {3.7.6},
  doi          = {10.5281/zenodo.591826},
  url          = {https://doi.org/10.5281/zenodo.591826}
}

@article{biessmann2021automated,
	title = {Automated data validation in machine learning systems},
	journal = {Bulletin of the IEEE Computer Society Technical Committee on Data Engineering},
	author = {Biessmann, Felix and Golebiowski, Jacek and Rukat, Tammo and Lange, Dustin and Schmidt, Philipp},
	year = {2021},
}

@article{schelterAutomatingLargeScaleData2018,
	title = {Automating {Large}-{Scale} {Data} {Quality} {Verification}},
	volume = {11},
	url = {https://doi.org/10.14778/3229863.3229867},
	doi = {10.14778/3229863.3229867},
	number = {12},
	journal = {PVLDB},
	author = {Schelter, Sebastian and Lange, Dustin and Schmidt, Philipp and Celikel, Meltem and Biessmann, Felix and Grafberger, Andreas and Ce-Likel, Meltem},
	year = {2018},
	pages = {1781--1794},
}

@misc{ISOIEC25024,
  author       = {ISO},
  title        = {Systems and software engineering --- Systems and software Quality Requirements and Evaluation (SQuaRE) --- Measurement of system and software product quality --- Data quality measurement},
  organization = {International Organization for Standardization and International Electrotechnical Commission},
  number       = {ISO/IEC 25024:2015},
  year         = {2015},
  address      = {Geneva},
  type         = {Standard},
  note         = {ISO/IEC Standard}
}

@article{ jung2025realistic,
  author  = {Philipp Jung and Sebastian J{\"a}ger and Nicholas Chandler and Felix Biessmann},
  title   = {Towards Realistic Error Models for Tabular Data},
  journal = {ACM Journal of Data and Information Quality},
  volume  = {17},
  number  = {4},
  pages   = {1--27},
  year    = {2025},
  doi     = {10.1145/3774914}
}

@InProceedings{jager2024conformal,
  title = 	 {From Data Imputation to Data Cleaning — Automated Cleaning of Tabular Data Improves Downstream Predictive Performance},
  author =       {J{\"a}ger, Sebastian and Biessmann, Felix},
  booktitle = 	 {Proceedings of The 27th International Conference on Artificial Intelligence and Statistics},
  pages = 	 {3394--3402},
  year = 	 {2024},
  volume = 	 {238},
  series = 	 {Proceedings of Machine Learning Research},
  publisher =    {PMLR},
  url =          {https://proceedings.mlr.press/v238/jager24a.html}
}

@article{jagerBenchmarkDataImputation2021,
  title = {A {{Benchmark}} for {{Data Imputation Methods}}},
  author = {J{\"a}ger, Sebastian and Allhorn, Arndt and Bie{\ss}mann, Felix},
  year = 2021,
  month = jul,
  journal = {Frontiers in Big Data},
  volume = {4},
  issn = {2624-909X},
  doi = {10.3389/fdata.2021.693674},
}

@inproceedings{jung2025mechdetect,
  author    ={Jung, Philipp and Chandler, Nicholas and J{\"a}ger, Sebastian and Biessmann, Felix},
  booktitle ={2025 International Conference on Data Science and Intelligent Systems (DSIS)}, 
  title     ={MechDetect: Detecting Data-Dependent Errors}, 
  year      ={2025},
  doi       ={10.1109/DSIS67228.2025.11390600}
}

@article{rubin1976inference,
  title={Inference and missing data},
  author={Rubin, Donald B.},
  journal={Biometrika},
  volume={63},
  number={3},
  pages={581--592},
  year={1976},
  publisher={Oxford University Press}
}

@article{jiang2025tabularsurvey,
         title={Representation Learning for Tabular Data: A Comprehensive Survey}, 
         author={Jun-Peng Jiang and
                 Si-Yang Liu and
                 Hao-Run Cai and
                 Qile Zhou and
                 Han-Jia Ye},
         journal={arXiv preprint arXiv:2504.16109},
         year={2025}
}

@article{abedjan2016detecting,
  author    = {Abedjan, Ziawasch and Chu, Xu and Deng, Dong and Fernandez, Raul Castro and Ilyas, Ihab F. and Ouzzani, Mourad and Papotti, Paolo and Stonebraker, Michael and Tang, Nan},
  title     = {Detecting Data Errors: Where are we and what needs to be done?},
  journal   = {Proceedings of the VLDB Endowment},
  volume    = {9},
  number    = {12},
  pages     = {993--1004},
  year      = {2016},
  doi       = {10.14778/2994509.2994518},
  publisher = {VLDB Endowment}
}

@article{powell2009impactOfErrors,
    author = {Powell, Stephen and Baker, Kenneth and Lawson, Barry},
    year = {2009},
    month = {05},
    pages = {126-132},
    title = {Impact of Errors in Operational Spreadsheets},
    volume = {47},
    journal = {Decision Support Systems},
    doi = {10.1016/j.dss.2009.02.002}
}

@book{silberschatz2019database,
  author    = {Silberschatz, Abraham and Korth, Henry F. and Sudarshan, S.},
  title     = {Database System Concepts},
  edition   = {7th},
  year      = {2019},
  publisher = {McGraw-Hill},
  address   = {New York, NY},
  isbn      = {9780078022159}
}

@article{bischl2021openml,
  title={{OpenML} Benchmarking Suites},
  author={Bischl, Bernd and Casalicchio, Giuseppe and Feurer, Matthias and Gijsbers, Pieter and Hutter, Frank and Lang, Michel and Mantovani, Rafael G. and van Rijn, Jan N. and Vanschoren, Joaquin},
  journal={Advances in Neural Information Processing Systems},
  volume={34},
  pages={23236--23247},
  year={2021}
}

@article{gurupur2025EHRerrors,
  title={Incompleteness of Electronic Health Records: An Impending Process Problem Within Healthcare},
  author={Gurupur, Varadraj and Hooshmand, Shadab and Prabhu, Deepthi F. and Trader, Emily and Salvi, Shreya},
  journal={Healthcare},
  volume={13},
  number={22},
  pages={2900},
  year={2025},
  month={Nov},
  day={13},
  doi={10.3390/healthcare13222900},
  pmid={41302288},
  pmcid={PMC12652376},
  publisher={MDPI}
}

@inproceedings{Prokhorenkova2018CatBoost,
  title     = {CatBoost: Unbiased Boosting with Categorical Features},
  author    = {Prokhorenkova, Liudmila and Gusev, Gleb and Vorobev, Aleksandr and Dorogush, Anna V. and Gulin, Andrey},
  booktitle = {Proceedings of the Thirty-Second Conference on Artificial Intelligence (AAAI)},
  year      = {2018},
  pages     = {6639--6647},
  url       = {https://ojs.aaai.org/index.php/AAAI/article/view/11746}
}

@inproceedings{Ke2017LightGBM,
  title     = {LightGBM: A Highly Efficient Gradient Boosting Decision Tree},
  author    = {Ke, Guolin and Meng, Qi and Finley, Thomas and Wang, Taifeng and Chen, Wei and Ma, Weidong and Ye, Qi and Liu, Tie-Yan},
  booktitle = {Advances in Neural Information Processing Systems 30 (NeurIPS)},
  year      = {2017},
  pages     = {3146--3154}
}

\end{document}